# PAIR: A Novel Large Language Model-Guided Selection Strategy for Evolutionary Algorithms

Shady Ali*[1], Mahmoud Ashraf [1], Seif Hegazy [1], Fatty Salem [1], Hoda Mokhtar [1], Mohamed Medhat Gaber [3], Mohamed Taher Alrefaie [1/2]

[1] Egypt University of Informatics, [2] Premio.AI, [3] Birmingham City University

*Corresponding author(s). E-mail: 22-101195@students.eui.edu.eg,
Contributing authors: 22-101201@students.eui.edu.eg, 22-101049@students.eui.edu.eg, fatty.salem@eui.edu.eg, hoda.mokhtar@eui.edu.eg, mohamed.gaber@bcu.ac.uk ,mohamed.taher@eui.edu.eg

## Abstract

Evolutionary Algorithms (EAs) employ random or simplistic selection methods, limiting their exploration of solution spaces and convergence to optimal solutions. The randomness in performing crossover or mutations may limit the model's ability to evolve efficiently. This paper introduces Preference-Aligned Individual Reciprocity (PAIR), a novel selection approach leveraging Large Language Models to emulate human-like mate selection, thereby introducing intelligence to the pairing process in EAs. PAIR prompts an LLM to evaluate individuals within a population based on genetic diversity, fitness level, and crossover compatibility, guiding more informed pairing decisions. We evaluated PAIR against a baseline method called LLM-driven EA (LMEA), published recently. Results indicate that PAIR significantly outperforms LMEA across various TSP instances, achieving lower optimality gaps and improved convergence. This performance is especially noticeable when combined with the flash thinking model, demonstrating increased population diversity to escape local optima. In general, PAIR provides a new strategy in the area of in-context learning for LLM-driven selection in EAs via sophisticated preference modelling, paving the way for improved solutions and further studies into LLM-guided optimization.

## 1 Introduction

The remarkable ability of Large Language Models (LLMs) to perform tasks through in-context learning has opened new avenues for their application across diverse domains. In-context learning, where models adapt to new tasks based on provided examples within a prompt, allows LLMs to generate high-quality text and code without requiring explicit fine-tuning (Brown et al., 2020a). This capability has spurred the exploration of LLMs in optimization tasks, where they act as intelligent guides steering search processes toward optimal solutions. This has led to what is called "Guided LLM Optimization," which leverages the in-context learning of LLMs to direct search-based algorithms to find optimal solutions in various problems.

Evolutionary Algorithms (EAs), a class of metaheuristic optimization algorithms, are effective at tackling complex problems by iteratively evolving a population of solutions (Eiben and Smith, 2015). EAs typically involve stages of selection, crossover, and mutation to refine the solution pool. However, conventional selection methods often rely on random or basic fitness-based selection, which can be limiting and hinder the exploration of diverse regions in the solution space. This limitation often leads to premature convergence and suboptimal solutions (Bhattacharya, 2014; Blickle and Thiele, 1996; Goldberg and Deb, 1991).

To address this, we introduce Preference-Aligned Individual Reciprocity (PAIR), a novel approach that integrates human-like decision-making into the selection phase of EAs. PAIR uses LLMs to evaluate the suitability of individuals for pairing, drawing inspiration from the way humans select partners, as done in dating apps. Specifically, PAIR prompts an LLM to analyze individuals in the population, considering their genetic diversity, fitness, and crossover compatibility. This informed pairing mechanism ensures a more strategic approach to the selection phase, fostering both the exploration of new solutions and the preservation of desirable traits. By bringing the reasoning power of LLMs directly into the selection process, PAIR not only leverages the search capability of EAs but also introduces an additional layer of intelligence to guide the overall optimization process, which can increase diversity in the population and lead to better solutions. We demonstrate the effectiveness of PAIR using the Traveling Salesman Problem (TSP), comparing it against LLM-driven EA (S. Liu et al., 2023), and evaluating the algorithm with various LLMs.

The remainder of this paper is structured as follows: Section 2 reviews related work, highlighting advancements in LLM-guided optimization and evolutionary computation. Section 3 details the methodology behind PAIR, including its implementation and integration with EAs. Section 4 presents experimental results, showcasing PAIR's performance on the Traveling Salesman Problem (TSP) across various instances. Finally, Section 5 discusses the broader implications of the findings, outlines limitations, and suggests directions for future research.

## 2 Related Work

The integration of advanced computational models into optimization frameworks has been a growing area of interest, with evolutionary algorithms (EAs) and large language models (LLMs) standing out as two distinct yet complementary paradigms (Guo et al., 2024; S. Liu et al., 2023; W. Liu et al., 2024). This section reviews the foundational advancements in these fields, their intersections, and the existing challenges that the proposed Preference-Aligned Individual Reciprocity (PAIR) approach addresses.

## 2.1 Introduction to Evolutionary Algorithms

Evolutionary algorithms have long been recognized for their capability to solve complex optimization problems by mimicking natural selection processes. Foundational works, such as (*Handb. Evol. Comput.*, 1997). These methods, while effective, face challenges in exploring diverse solution spaces and avoiding premature convergence.

Traditional selection methods, as analyzed by (Blickle and Thiele, 1996) and (Goldberg and Deb, 1991), often rely on fitness-proportionate or rank-based techniques. While these approaches maintain a balance between exploration and exploitation, they can struggle with scalability and diversity retention in dynamic environments. These limitations have spurred research into more sophisticated selection mechanisms, paving the way for intelligent integration with machine learning models (Eiben and Smith, 2015; Karafotias et al., 2015).

## 2.2 In-Context Learning and LLM Applications

The emergence of in-context learning has revolutionized how large language models adapt to various tasks without explicit retraining. The seminal work of (Brown et al., 2020b) introduced GPT-3, showcasing its few-shot learning capabilities and potential for guided decision-making. Subsequent advancements, such as the unified pre-training framework by (Dong et al., 2019), further enhanced model adaptability by integrating understanding and generation tasks.

Leveraging LLMs for metaheuristics and decision-making has opened new avenues for optimization. (Wei et al., 2022) demonstrated the efficacy of chain-of-thought prompting in enabling complex reasoning, while (Kojima et al., 2022) highlighted the zero-shot reasoning capabilities of LLMs in solving novel tasks. Despite these advancements, studies like those by (Zhao et al., 2024) underline the limitations in model adaptability and performance consistency, particularly in optimization contexts.

OPRO, by Yang et al. (2023), investigated doing Optimization by Prompting, where they leveraged LLMs as optimizers that given the problem statement and previous solutions, generate new solutions. These solutions are then iteratively evaluated and improved as the process continues. They mainly focused on using OPRO for prompt optimization, where they aimed to optimize the prompt fed to the LLM to enhance the quality of generated results and solutions.

## 2.3 Advancements in EAs and LLMs

Traditional evolutionary algorithms have seen significant enhancements through the development of advanced selection schemes. Blickle and Thiele, (1996) provided a mathematical analysis of tournament selection, highlighting its efficiency in maintaining diversity. Similarly, Miller and Goldberg, (1995) explored the robustness of selection mechanisms in noisy environments, emphasizing their impact on convergence reliability.

Among these advancements, the Large Language Model-driven Evolutionary Algorithm (LMEA) framework, introduced by (Y. Liu et al., 2023). S. Liu et al., (2023) represents a pivotal integration of LLMs into evolutionary processes. LMEA employs LLMs to execute key genetic operations, including selection, crossover, and mutation, by generating offspring based on textual prompts describing the evolutionary state. This framework leverages the in-context learning capabilities of LLMs to adaptively adjust evolutionary strategies, such as dynamically modifying temperature parameters to balance exploration and exploitation. Empirical evaluations of LMEA on Traveling Salesman Problem (TSP) instances demonstrated competitive performance with traditional heuristics, underscoring its potential to reduce reliance on domain-specific operators and enhance solution diversity (Y. Liu et al., 2023).

Intelligent integration into EAs has been a focus of research, with (Stanhope and Daida, 1998) (Grefenstette, 1986) investigating the optimization of control parameters for dynamic environments. These studies set the stage for hybrid approaches that incorporate machine learning techniques to address the inherent limitations of traditional methods.

Traditional evolutionary algorithms have seen significant enhancements through the development of advanced selection schemes. (Blickle and Thiele, 1996) provided a mathematical analysis of tournament selection, highlighting its efficiency in maintaining diversity. Similarly, Miller and Goldberg, (1995) explored the robustness of selection mechanisms in noisy environments, emphasizing their impact on convergence reliability.

Intelligent integration into EAs has been a focus of research, with Stanhope and Daida, (1998) and Grefenstette, (1986) investigating the optimization of control parameters for dynamic environments. These studies set the stage for hybrid approaches that incorporate machine learning techniques to address the inherent limitations of traditional methods.

### 2.4 Hybrid Optimization Frameworks

Hybrid frameworks combining machine learning and traditional optimization have proven effective in tackling complex problems. Zhou, (2018) explored weakly supervised learning's role in optimization, while Bengio et al., (2021) provided a methodological overview of machine learning's application in combinatorial tasks. (Zhang and Li, 2007) MOEA/D framework demonstrated the efficacy of decomposition-based multi-objective optimization, and (Talbi, 2002) taxonomy of hybrid metaheuristics analyzed their strengths and limitations in diverse scenarios.

### 2.5 The Research Gap

The paper identified a research gap in the relatively new domain in identifying an LLM-driven approach in their selection mechanisms. In general, existing frameworks integrating large language models (LLMs) with evolutionary algorithms have demonstrated promising advancements, particularly in leveraging LLMs' reasoning capabilities to enhance evolutionary operations. For instance, Liu et al.'s (2023) introduced the LMEA framework, which incorporated LLMs for performing evolutionary tasks such as crossover and mutation. However, current methods largely emphasize operational enhancements while paying limited attention to selection mechanisms that align with individual preferences and adaptive diversity. This gap highlights the need for a more focused approach that leverages LLMs to refine selection processes, ensuring both diversity and convergence are effectively balanced in optimization.

## 3 Methodology

This section details the proposed methodology, Preference-Aligned Individual Reciprocity (PAIR), for the selection phase of Evolutionary Algorithms (EAs). PAIR leverages Large Language Models (LLMs) to introduce a preference-based selection process akin to human mate selection that is based on mutual interest between the two individuals. The implementation of PAIR consists of three primary stages: Initialization, PAIR-driven Selection, and Population Evolution, which are iteratively applied until a satisfactory solution is achieved. We also use the LLM to utilize and execute evolutionary operators, and we adopted the LLM Adaptive Temperature mechanism from LMEA's approach (S. Liu et al., 2023).

### 3.1 Initialization

Initially, the algorithm generates a population of $N$ individuals randomly, see Fig. 1. Each individual represents a potential solution to the problem, in our case the Traveling Salesman Problem (TSP), and is encoded as a sequence of nodes that defines a route. In each iteration, the algorithm will then send the current population alongside their solution's lengths to the LLM, to perform the selection, crossover, and mutation operators.

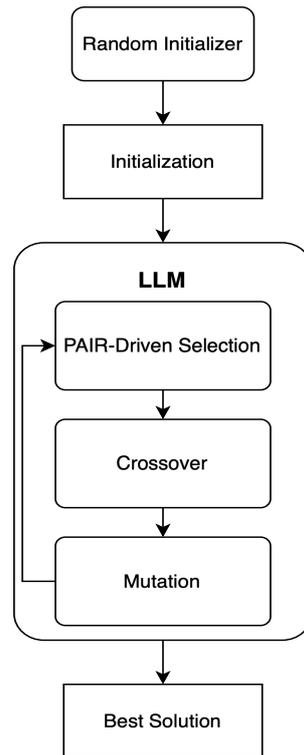

Figure 1: PAIR's Architecture and flow, initialize a population of candidate solutions randomly, then the LLM guides the population generation loop starting with the PAIR-Driven selection, then cross over and mutation. The LLM uses the mutated individuals again for PAIR Selection in the same iteration they were produced in as PAIR Selection cannot use the same individual twice.

| Approach | Initialization | Selection | LLM Utilization | Adaptive LLM Temperature |
|---|---|---|---|---|
| *OPRO* | Random | ✗ | Creates new solutions iteratively different from previous generation's solutions | ✗ |
| *LMEA* | Random | Random | Uses and excutes EA operators. | ✓ |
| ***PAIR*** | Random | **PAIR-Driven Selection** | Uses and excutes EA operators | ✓ |

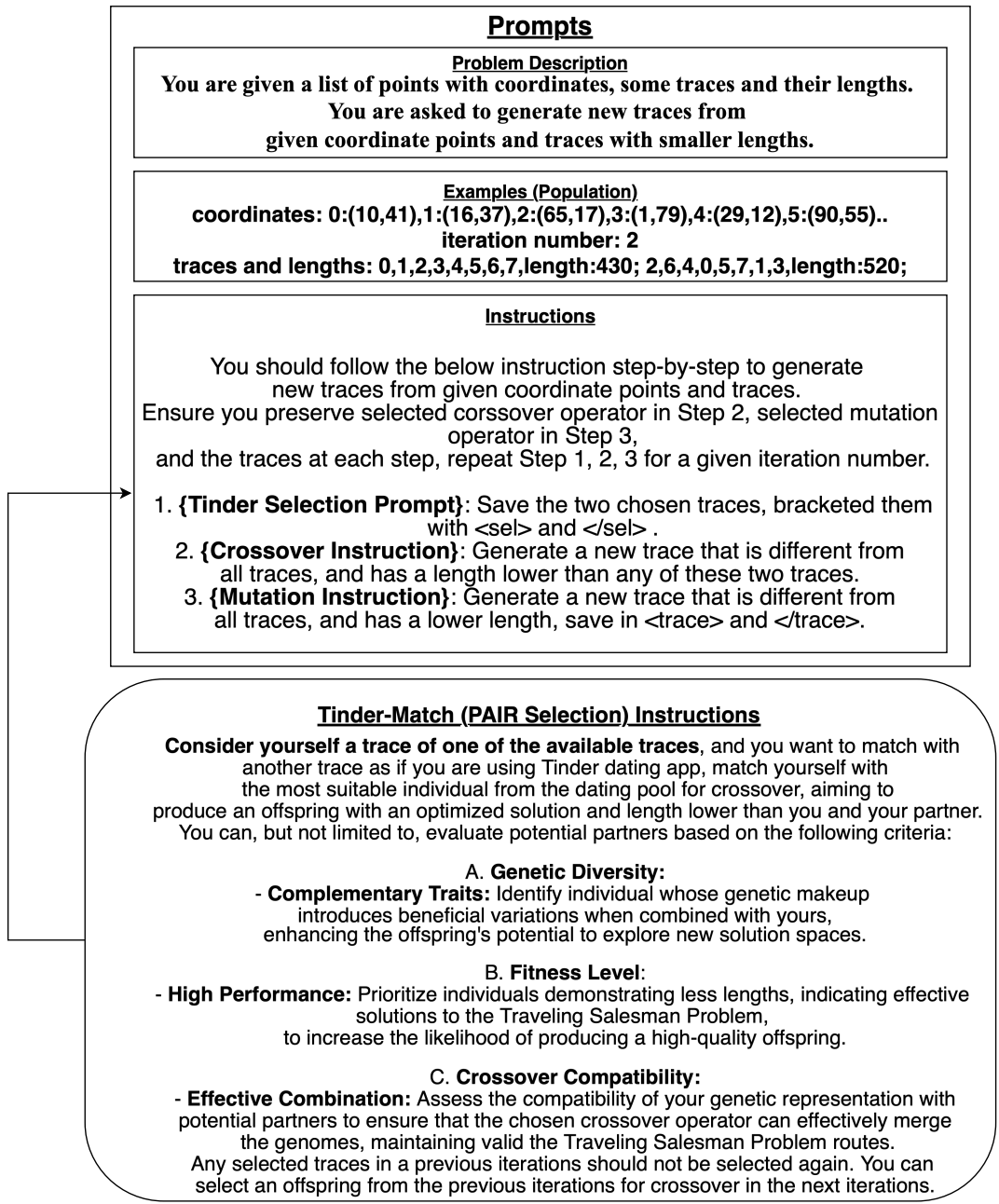

Figure 3: Prompts and instructions used for the problem statement, examples, and the intructions for the LLM to follow. The first text defines the problem statement and description to provide context to the model. Then the second text provides some examples for population, the number of iterations (how many individuals to produce in this generation), and the solution of each individual from the previous population. This prompt is the one sent to the LLM as a user prompt for each generation. The last text provides the instructions for the LLM on what steps to do (selection, crossover, mutation), and what Evolutionary Operators to use in the process. The first step is the selection which is PAIR-Driven selection as provided in the lower box. And step 2 and 3 for the provided Crossover and mutation operators and how to use them.

### 3.2 PAIR-Driven Selection

The core novelty of this methodology lies in the selection process, which is governed by the PAIR approach. In this step, an LLM is prompted to act as an evaluator, simulating the preferences one might employ in a dating scenario. The LLM receives a population of individuals (traces of solution routes and their lengths), where each trace corresponds to a single solution, and the prompt instructs it to evaluate and select pairs of individuals for crossover. To simulate a human-like selection process, the LLM is instructed to first choose an individual, then, like role-playing, consider itself as that individual and choose the best, and most compatible individual from the rest of the population, given specific criteria, like how humans select their partners based on their 'standards'. The selection is guided by three distinct yet interrelated criteria: (1) genetic diversity: individuals are selected such that their traits offer complementary variations when combined, maximizing the exploration of new solution spaces; (2) fitness level: the pairing emphasizes individuals with shorter tour lengths, indicating superior performance; and (3) crossover compatibility: the selected pairings should have genetic structures that can combine to produce valid offspring routes using the selected crossover operator, ensuring effective genetic exchange. See Fig. 2 for a comparison between LMEA and the proposed PAIR algorithm.

Mirroring human monogamy, once two traces are selected, they are removed from the selection pool for the current generation (selection without replacement). To address the issue of the LLM potentially having too few options for later generations when half the population is already paired up, the offspring of earlier iterations are added to the selection pool for consideration, giving the LLM more options. The result of this stage is the generation of matched pairs for the next phase of the algorithm. See Fig. 3 for detailed step by step on the prompts and instructions given to the model.

### 3.3 Population Evolution

Following the selection stage, matched pairs are then used in crossover and mutation stages. A Crossover operator, which is also dictated by the LLM, is then performed on the selected pairs, and a new offspring is produced by combining the genetic material from two selected traces, ensuring the new offspring is a valid solution by using crossover operators that preserve valid solutions in the TSP. A Mutation operator is used to introduce variability in the offspring population, the selection of this mutation operator is guided by the LLM as well. The result of this stage is a population of new traces, which are then combined with the current population and sorted based on their fitness. Only the top N solutions are then kept for the next iteration, as stated by the classic Evolutionary Algorithm approach. This process is repeated for a fixed number of generations until an optimal or sufficiently good solution is found. We also adopt the Adaptive Temperature mechanism used by LMEA, where the LLM's sampling temperature is increase by 0.05 every 20 generations with no improvement, with a maximum temperature of 2.0.

### 3.4 Algorithmic Implementation and Evaluation

The PAIR methodology is implemented using *Gemini 2.0 Flash Experimental* and *Flash Thinking Experimental* LLMs (Google et al., 2024) to facilitate and drive the selection, crossover and mutation steps. We evaluate the performance of this approach on the Traveling Salesman Problem (TSP) using 2-Dimensional Euclidean TSP instances. The test cases include 4 node counts (10, 15, 20, 25) with 5 problem instances created by placing nodes randomly ("rue" instances), and 5 problem instances generated by clustering nodes around centers ("clu" instances). Results are measured in terms of optimality gap and convergence speed compared to a baseline approach: the LMEA algorithm. By evaluating our approach against these metrics, we aim to demonstrate the efficacy of PAIR-driven selection in improving the solution quality and the overall efficiency of EAs.

We found that comparing PAIR to LMEA's approach uniquely would be more insightful for the following reasons:

1. In the early stages of experimentation, we found that LMEA's approach yielded significantly different results depending on the LLM used, especially when compared with the results reported by the authors of LMEA using GPT-3.5 Turbo (Y. Liu et al., 2023).
2. LMEA's authors (Y. Liu et al., 2023) demonstrated that their approach outperforms the most recent state-of-the-art approach using LLMs in Evolutionary Algorithms, OPRO (Yang et al., 2023), in addition to achieving performance competitive to traditional heuristics on TSP instances.
3. PAIR is more comparable to LMEA since both approaches use LLMs to execute Evolutionary Operations, unlike approaches such as OPRO where the LLM directly generates new populations iteratively based on the previous generation.

Therefore, we decided to limit our direct comparison to LMEA only, given the LLMs computation constraints, the need to rerun all experiments to maintain accuracy and fairness in the comparisons, and that LMEA already demonstrated state-of-the-art performance compared to other powerful traditional heuristics like NN (Nearest Neighbor) and LLM-based approaches like OPRO.

## 4 Results

We demonstrate our results by calculating the optimality gap between the best solution found and optimal solution:

$$OptimalityGap = \frac{\text{Distance}_{\text{Found}} - \text{Distance}_{\text{Optimal}}}{\text{Distance}_{\text{Optimal}}}$$

In the results' tables, we average the optimality gap of 5 problems of each node count-problem type pair for all 8 problems:

$$OptimalityGap_{mean} = \sum_{i=1}^{5} \frac{OptimalityGap_i}{5}$$

To avoid bias in comparison, we used the same population size (P=16), maximum number of generations (250), and LLM (Gemini 2.0 Flash Experimental) when testing both approaches. We also used the same adaptive-temperature approach introduced in LMEA's paper in both experiments to make the selection methodology the only variable aspect.

Table 1: Comparison between LMEA and PAIR using Gemini 2.0 Flash Experimental. NaN means Not a Number, Not applicable in other terms.

| | Optimality Gap (%) ($Mean \pm$ Standard Deviation) | | Success Step (# Generations) ($Mean \pm$ Standard Deviation) | |
|---|---|---|---|---|
| **Problem** | LMEA | PAIR | LMEA | PAIR |
| rue-10 | **0.00 $\pm$ 0.00** | 5.88 $\pm$ 5.45 | **76.20 $\pm$ 78.90** | 146.0 $\pm$ *NaN* |
| rue-15 | 31.55 $\pm$ 13.36 | **26.7 $\pm$ 13.9** | *NaN $\pm$ NaN* | *NaN $\pm$ NaN* |
| rue-20 | 78.18 $\pm$ 10.45 | **57.08 $\pm$ 20.26** | *NaN $\pm$ NaN* | *NaN $\pm$ NaN* |
| rue-25 | 107.72 $\pm$ 26.16 | **81.01 $\pm$ 9.9** | *NaN $\pm$ NaN* | *NaN $\pm$ NaN* |
| clu-10 | 1.76 $\pm$ 3.94 | **2.03 $\pm$ 2.9** | 90.75 $\pm$ 100.25 | **40.67 $\pm$ 29.87** |
| clu-15 | 26.25 $\pm$ 11.82 | **16.26 $\pm$ 10.35** | *NaN $\pm$ NaN* | *NaN $\pm$ NaN* |
| clu-20 | **51.89 $\pm$ 27.11** | 53.93 $\pm$ 21.44 | *NaN $\pm$ NaN* | *NaN $\pm$ NaN* |
| clu-25 | 114.87 $\pm$ 19.25 | **105.49 $\pm$ 21.23** | *NaN $\pm$ NaN* | *NaN $\pm$ NaN* |

In results Table 1, comparing the results based on average optimality gap, the PAIR approach outperforms the random selection approach used by LMEA in 5 of 8 problem types. In some problems, e.g. clu-15, applying the PAIR selection resulted in 38% or more improvement in results over standard LMEA.

Table 2: Demonstrating PAIR using Gemini 2.0 Flash Thinking Experimental. NaN means Not a Number, Not applicable in other terms.

| | Optimality Gap (%) ($Mean \pm$ Standard Deviation) | | | Success Step (# Generations) ($Mean \pm$ Standard Deviation) | | |
|---|---|---|---|---|---|---|
| | LMEA | PAIR | PAIR | LMEA | PAIR | PAIR |
| Problem | Gemini 2.0 Flash | Gemini 2.0 Flash | Gemini 2.0 Flash Thinking | Gemini 2.0 Flash | Gemini 2.0 Flash | Gemini 2.0 Flash Thinking |
| rue-10 | **0.00 $\pm$ 0.00** | 5.88 $\pm$ 5.45 | **0.00 $\pm$ 0.00** | 76.20 $\pm$ 78.90 | 146.0 $\pm$ *NaN* | **75.6 $\pm$ 61.79** |
| rue-15 | 31.55 $\pm$ 13.36 | 26.7 $\pm$ 13.9 | **11.32 $\pm$ 9.07** | *NaN $\pm$ NaN* | *NaN $\pm$ NaN* | **39.0 $\pm$ NaN** |
| rue-20 | 78.18 $\pm$ 10.45 | 57.08 $\pm$ 20.26 | **33.57 $\pm$ 20.85** | *NaN $\pm$ NaN* | *NaN $\pm$ NaN* | *NaN $\pm$ NaN* |
| rue-25 | 107.72 $\pm$ 26.16 | 81.01 $\pm$ 9.9 | **61.07 $\pm$ 9.87** | *NaN $\pm$ NaN* | *NaN $\pm$ NaN* | *NaN $\pm$ NaN* |
| clu-10 | 1.76 $\pm$ 3.94 | 2.03 $\pm$ 2.9 | **0.00 $\pm$ 0.00** | 90.75 $\pm$ 100.25 | **40.67 $\pm$ 29.87** | 46.0 $\pm$ 38.34 |
| clu-15 | 26.25 $\pm$ 11.82 | 16.26 $\pm$ 10.35 | **14.87 $\pm$ 7.78** | *NaN $\pm$ NaN* | *NaN $\pm$ NaN* | *NaN $\pm$ NaN* |
| clu-20 | 51.89 $\pm$ 27.11 | 53.93 $\pm$ 21.44 | **21.93 $\pm$ 6.67** | *NaN $\pm$ NaN* | *NaN $\pm$ NaN* | *NaN $\pm$ NaN* |
| clu-25 | 114.87 $\pm$ 19.25 | 105.49 $\pm$ 21.23 | **64.93 $\pm$ 17.31** | *NaN $\pm$ NaN* | *NaN $\pm$ NaN* | *NaN $\pm$ NaN* |

In results Table 2, we tried using PAIR with Gemini 2.0 Flash Thinking Experimental as the LLM to investigate how the approach performs with more advanced models, and it resulted in significantly better results. The combination of PAIR and Gemini 2.0 flash thinking outperformed the previous two experiments in all 8 problems with improvement margins reaching 57.7% in clu-20, for example.

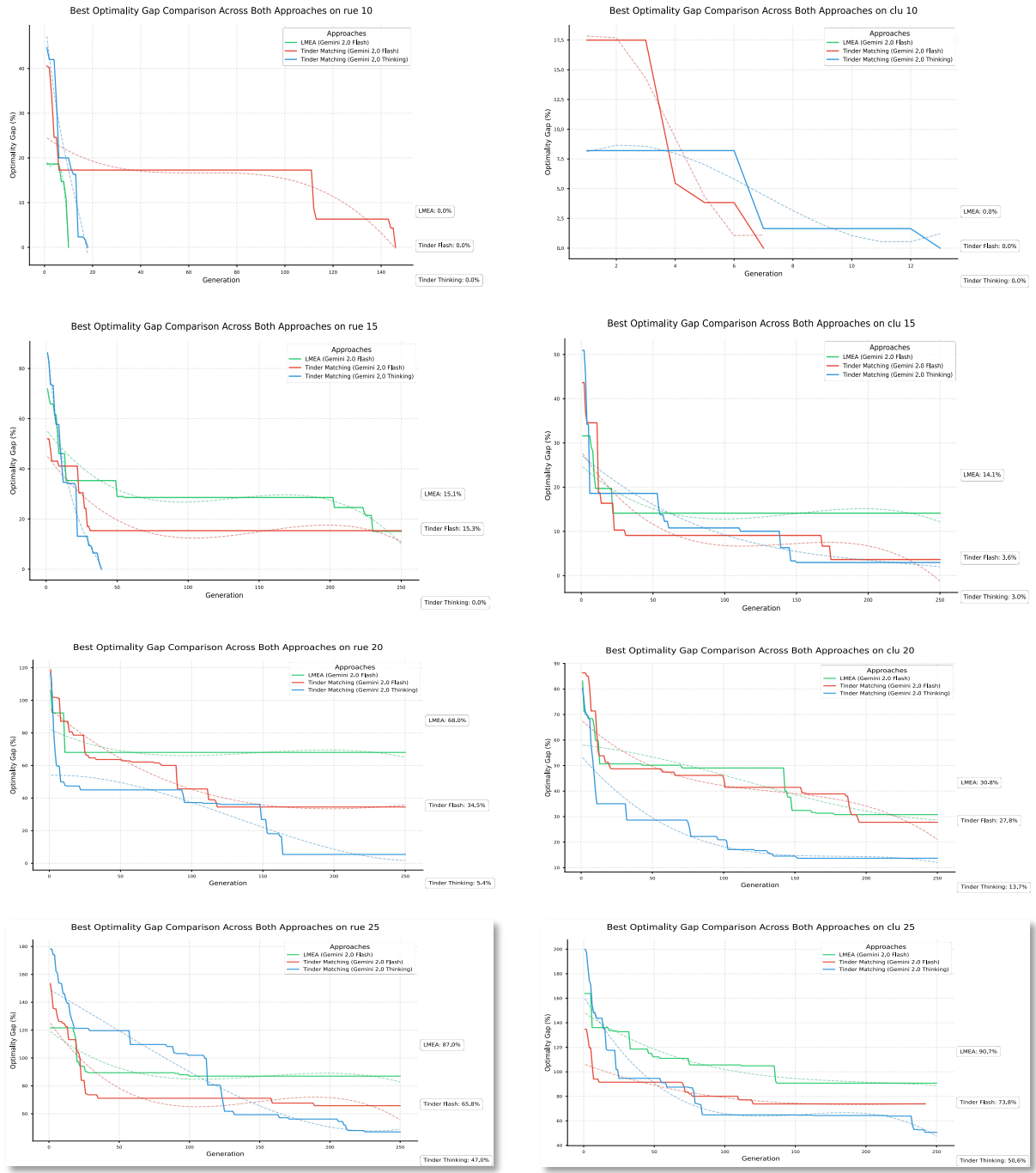

*Figure 4: Comparison of convergence between LMEA (Gemini 2.0 Flash Experimental) and PAIR (Gemini 2.0 Flash Experimental and Flash Thinking Experimental). The graphs illustrate the progression of the optimality gap (on y-axis) achieved by each approach (each represented by a colored line) over successive generations for the eight problems tested.*

In Fig 4, we compared the best performance of each approach-llm pair across each problem type-node count pairs. Specifically, we picked the best performance of each approach in the 5 problems of each set (rue-10, clu-10, etc.) to compare the difference between them in their convergence pattern. All graphs have the convergence of LMEA on Gemini 2.0 Flash and PAIR on both Gemini 2.0 Flash, and Gemini 2.0 Flash Thinking.

Focusing on the two approaches with Gemini 2.0 Flash (LMEA: Green, PAIR: Red), LMEA converged earlier in both 10 Node problems (rue-10 & clu-10, from initialization in clu-10), PAIR outperformed LMEA in almost all other problems, as LMEA got stuck several times in sub-optimal solutions while PAIR converged better at them. For PAIR on Gemini 2.0 Flash Thinking, it outperformed them both whether in finding the best optimality gap or the convergence pattern, especially as the node count increased.

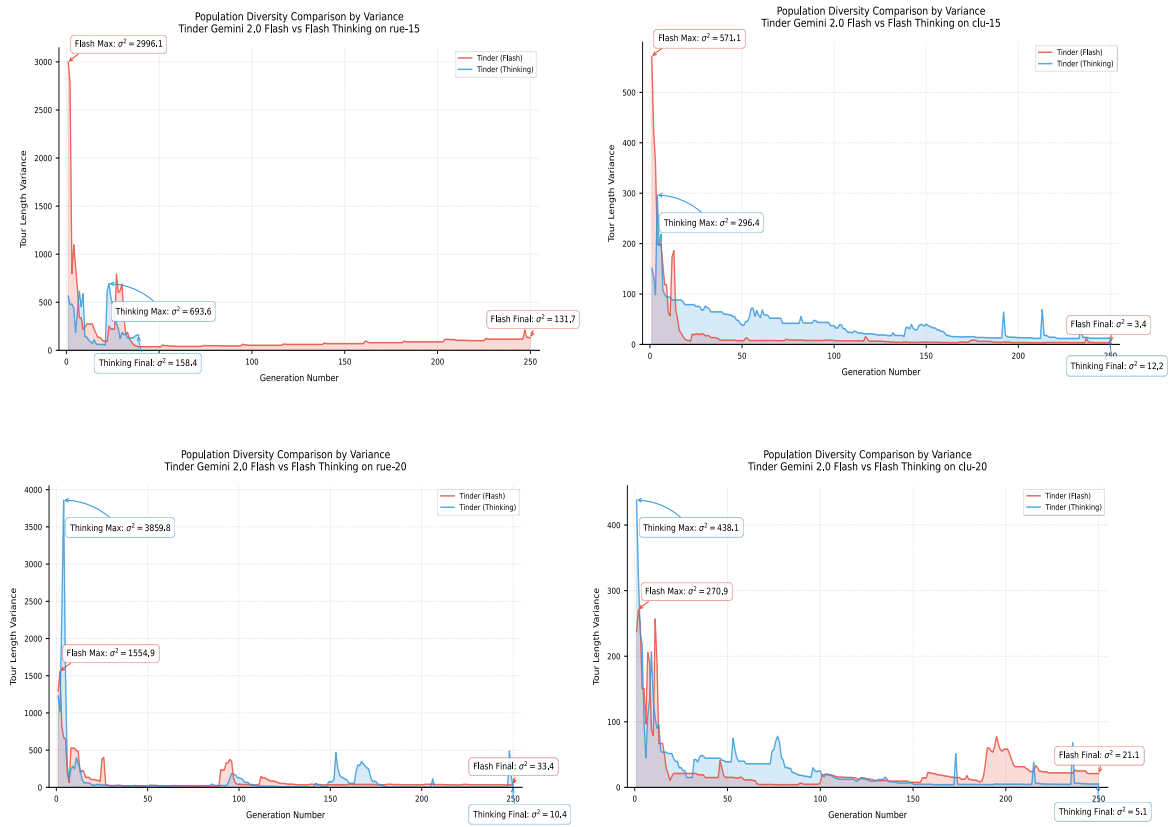

*Figure 5: Comparison of Population Diversity in best performance problems between PAIR (Gemini 2.0 Flash Experimental) and PAIR (Gemini 2.0 Flash Thinking Experimental). The graphs show variance of solution (tour) lengths in a population (plotted on the y-axes) over successive generations (plotted on the x-axes). Tour lengths population variance is used as a measure of population diversity which can influence the algorithms' tendency to explore different, potentially suboptimal, solutions, and exploiting best solutions already discovered.*

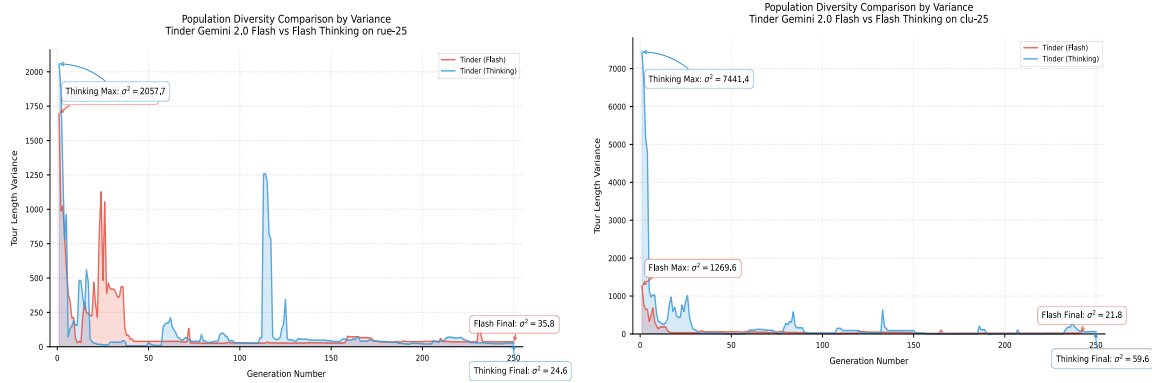

In Fig. 5, we tried to visualize the diversity of the population across generations for the best performance problem instance in all problem sets except 10 Node problems (rue-10 & clu-10) on our approach using both Gemini 2.0 Experimental models (Flash & Flash Thinking). We measured the diversity using the Variance of the tour lengths of each population.

One pattern we did notice is the populations generated by the Thinking model always had higher variance (more diverse) across generations. This pattern suggests that having more diverse population can help the algorithm escape more sub-optimal solutions and converge better over generations and that PAIR on Gemini 2.0 Flash Thinking was able to maintain more diverse population across generations. We could then infer that reasoning models are potentially better at PAIR-Based Selection on local level (population diversity) and global level (final solutions). Code and used data of this study are publicly released on Github[1].

# Discussion

This study introduced Preference-Aligned Individual Reciprocity (PAIR), a novel selection methodology for Evolutionary Algorithms (EAs) leveraging the in-context learning capabilities of Large Language Models (LLMs). The primary objective was to evaluate the efficacy of PAIR in comparison to the established LMEA baseline (S. Liu et al., 2023), particularly in the context of solving the Traveling Salesman Problem (TSP). The ensuing discussion interprets the key findings, contextualizes them within existing research, addresses the inherent limitations of the current study, and proposes pertinent future research directions.

The experimental results demonstrated the enhanced performance of PAIR over the LMEA baseline across a spectrum of TSP instances. Notably, PAIR exhibited a statistically significant reduction in the optimality gap compared to LMEA. For instance, in the clu-15 problem set, PAIR achieved an average optimality gap of 16%, representing a 38% improvement over LMEA's 26%. This enhanced solution quality can be attributed to PAIR's intelligent selection mechanism, which, unlike LMEA's random selection, strategically pairs individuals based on genetic diversity, fitness level, and crossover compatibility.

Furthermore, the convergence analysis revealed that PAIR facilitated a more rapid convergence towards superior solutions, mitigating the propensity for stagnation in suboptimal regions of the search space often observed with LMEA. The accelerated convergence suggests a more efficient exploration-exploitation trade-off facilitated by the LLM-driven preference alignment.

---

[1] https://github.com/SHIXOOM/PAIR

The differential performance observed between the Gemini 2.0 Flash Experimental and Flash Thinking models within the PAIR framework was obvious in the results due to several reasons. First, it underscores the importance of the LLM's reasoning capabilities in effectively modeling the selection preferences. While both PAIR implementations outperformed LMEA in several instances, the magnitude of improvement was more pronounced with the Flash Thinking model, indicating a positive correlation between the sophistication of the LLM and the algorithmic efficacy of PAIR.

Analysis of population diversity further demonstrates reasoning LLMs potential. The consistently higher population variance observed with Gemini 2.0 Flash Thinking, in contrast with Gemini 2.0 Flash, suggests a greater capacity for exploring diverse regions of the solution space, leading to higher convergence rate, better final solutions and efficient navigation through sub-optimum solutions.

Second, this observation has implications for the selection of appropriate LLMs in guided optimization strategies, in general. The foundational principle of PAIR, drawing inspiration from human mate selection, provides an intuitive yet effective heuristic for guiding the selection process in EAs. Unlike the agnostic selection mechanism inherent in LMEA, PAIR introduces a layer of informed decision-making, demonstrably leading to enhanced optimization outcomes.

On the other hand, the findings of this study have also proved that LLM-guided optimization can be used to solve EA problems better than traditional methods. While existing research has explored the utility of LLMs in various optimization tasks, PAIR distinguishes itself by specifically focusing on the selection phase within EAs, offering a departure from conventional methods. In comparison to LMEA, which relies on a random selection strategy, PAIR introduces a level of strategic decision-making previously absent. This aligns with the broader trend in evolutionary computation towards incorporating more informed and adaptive selection mechanisms. Furthermore, the performance gains observed with PAIR resonate with findings in studies exploring diversity maintenance strategies in EAs (Guo et al., 2024; S. Liu et al., 2024; Wu et al., 2024) . The LLM's ability to assess genetic diversity and promote pairings that enhance population variance aligns with established principles for avoiding premature convergence and fostering robust exploration.

Despite the promising results, this study is subject to certain limitations. The computational cost associated with invoking LLMs for each selection step represents a potential practical constraint, particularly when compared to the computationally inexpensive random selection of LMEA, especially when some studies reported that code-based models outperform generic models in structured thinking and analytics (Alrefaie et al., 2025). Future research should explore strategies for optimizing the efficiency of PAIR, potentially through batch processing or more lightweight LLM implementations. The generalizability of PAIR to optimization problems beyond the TSP warrants further investigation. While the principles of preference-aligned selection may be broadly applicable, empirical validation across diverse problem domains is necessary.

Furthermore, exploring the sensitivity of PAIR to different prompting strategies and the inherent biases within the employed LLMs constitutes a crucial gap for future studies. In addition, investigating alternative or augmented selection criteria for the LLMs might lead to further performance enhancements. Exploring hybrid approaches that selectively apply PAIR based on the stage of the evolutionary process or population characteristics could also offer a promising avenue for balancing efficacy and computational cost.

## 5 Conclusion and Future Work

In conclusion, this study introduced and rigorously evaluated Preference-Aligned Individual Reciprocity (PAIR), a novel selection methodology for Evolutionary Algorithms (EAs) that leverages the in-context learning capabilities of Large Language Models (LLMs). By emulating human-like preference modelling, PAIR demonstrably enhances the selection process, leading to significant performance gains compared to the LMEA baseline on the Traveling Salesman Problem (TSP). The intelligent selection mechanism inherent in PAIR facilitates improved solution quality, as evidenced by the statistically significant reduction in optimality gaps across various TSP instances. Furthermore, PAIR promotes accelerated convergence towards superior solutions and fosters enhanced population diversity, indicating a more effective exploration-exploitation trade-off compared to the random selection strategy employed by LMEA. The superior performance observed when utilizing the Gemini 2.0 Flash Thinking model underscores the importance of the LLM's reasoning capabilities in effectively guiding the evolutionary search.

While these findings are promising, they also highlight avenues for future research. A primary direction involves mitigating the inherent computational cost associated with LLM invocations within the selection loop. Investigations into techniques such as batch processing of selection candidates or the exploration of more computationally efficient LLM architectures are warranted. Expanding the empirical evaluation of PAIR to a broader range of optimization problems beyond the TSP is crucial to ascertain the generalizability of its benefits. Future studies should also explore the sensitivity of PAIR's performance to variations in prompting strategies and explore the potential biases embedded within different LLMs. Furthermore, the exploration of alternative or supplementary selection criteria for the LLM to consider offers a promising avenue for further performance optimization. From a theoretical standpoint, a rigorous analysis of PAIR's convergence properties, comparing them to established EA frameworks, would provide valuable insights into its long-term behavior and effectiveness. Finally, the development of hybrid approaches that strategically combine PAIR with other advanced EA techniques could potentially leverage the strengths of both methodologies, offering a balanced approach to optimization. These future endeavors will serve to further solidify the theoretical underpinnings and practical applicability of PAIR as a valuable contribution to the field of evolutionary computation and LLM-guided optimization.

H., Barr, I., Miao, Y., Natsev, P., Devlin, J., Behbahani, F., Prost, F., Sun, Yanhua, Myaskovsky, A., Pillai, T.S., Hurt, D., Lazaridou, A., Xiong, Xi, Zheng, C., Pardo, F., Li, Xiaowei, Horgan, D., Stanton, J., Ambar, M., Xia, F., Lince, A., Wang, M., Mustafa, B., Webson, A., Lee, H., Anil, R., Wicke, M., Dozat, T., Sinha, Abhishek, Piqueras, E., Dabir, E., Upadhyay, S., Boral, A., Hendricks, L.A., Fry, C., Djolonga, J., Su, Y., Walker, J., Labanowski, J., Huang, R., Misra, V., Chen, Jeremy, Skerry-Ryan, R., Singh, A., Rijhwani, S., Yu, D., Castro-Ros, A., Changpinyo, B., Datta, R., Bagri, S., Hrafnkelsson, A.M., Maggioni, M., Zheng, D., Sulsky, Y., Hou, S., Paine, T. Le, Yang, A., Riesa, J., Rogozinska, D., Marcus, D., Badawy, D. El, Zhang, Q., Wang, Luyu, Miller, H., Greer, J., Sjos, L.L., Nova, A., Zen, H., Chaabouni, R., Rosca, M., Jiang, J., Chen, C., Liu, Ruibo, Sainath, T., Krikun, M., Polozov, A., Lespiau, J.-B., Newlan, J., Cankara, Zeyncep, Kwak, S., Xu, Yunhan, Chen, P., Coenen, A., Meyer, C., Tsihlas, K., Ma, A., Gottweis, J., Xing, J., Gu, C., Miao, J., Frank, C., Cankara, Zeynep, Ganapathy, S., Dasgupta, I., Hughes-Fitt, S., Chen, H., Reid, D., Rong, K., Fan, H., van Amersfoort, J., Zhuang, V., Cohen, A., Gu, S.S., Mohananey, A., Ilic, A., Tobin, T., Wieting, J., Bortsova, A., Thacker, P., Wang, E., Caveness, E., Chiu, J., Sezener, E., Kaskasoli, A., Baker, S., Millican, K., Elhawaty, M., Aisopos, K., Lebsack, C., Byrd, N., Dai, H., Jia, W., Wiethoff, M., Davoodi, E., Weston, A., Yagati, L., Ahuja, A., Gao, I., Pundak, G., Zhang, S., Azzam, M., Sim, K.C., Caelles, S., Keeling, J., Sharma, A., Swing, A., Li, YaGuang, Liu, C., Bostock, C.G., Bansal, Y., Nado, Z., Anand, A., Lipschultz, J., Karmarkar, A., Proleev, L., Ittycheriah, A., Yeganeh, S.H., Polovets, G., Faust, A., Sun, J., Rrustemi, A., Li, P., Shivanna, R., Liu, Jeremiah, Welty, C., Lebron, F., Baddepudi, A., Krause, S., Parisotto, E., Soricut, R., Xu, Z., Bloxwich, D., Johnson, Melvin, Neyshabur, B., Mao-Jones, J., Wang, Renshen, Ramasesh, V., Abbas, Z., Guez, A., Segal, C., Nguyen, D.D., Svensson, J., Hou, L., York, S., Milan, K., Bridgers, S., Gworek, W., Tagliasacchi, M., Lee-Thorp, J., Chang, M., Guseynov, A., Hartman, A.J., Kwong, M., Zhao, R., Kashem, S., Cole, E., Miech, A., Tanburn, R., Phuong, M., Pavetic, F., Cevey, S., Comanescu, R., Ives, R., Yang, S., Du, C., Li, B., Zhang, Zizhao, Iinuma, M., Hu, C.H., Roy, A., Bijwadia, S., Zhu, Z., Martins, D., Saputro, R., Gergely, A., Zheng, S., Jia, D., Antonoglou, I., Sadovsky, A., Gu, S., Bi, Y., Andreev, A., Samangooei, S., Khan, M., Kocisky, T., Filos, A., Kumar, C., Bishop, C., Yu, A., Hodkinson, S., Mittal, S., Shah, P., Moufarek, A., Cheng, Yong, Bloniarz, A., Lee, Jaehoon, Pejman, P., Michel, P., Spencer, S., Feinberg, V., Xiong, Xuehan, Savinov, N., Smith, Charlotte, Shakeri, S., Tran, D., Chesus, M., Bohnet, B., Tucker, G., von Glehn, T., Muir, C., Mao, Y., Kazawa, H., Slone, A., Soparkar, K., Shrivastava, D., Cobon-Kerr, J., Sharman, M., Pavagadhi, J., Araya, C., Misiunas, K., Ghelani, N., Laskin, M., Barker, D., Li, Q., Briukhov, A., Houlsby, N., Glaese, M., Lakshminarayanan, B., Schucher, N., Tang, Y., Collins, E., Lim, H., Feng, F., Recasens, A., Lai, G., Magni, A., De Cao, N., Siddhant, A., Ashwood, Z., Orbay, J., Dehghani, M., Brennan, J., He, Y., Xu, K., Gao, Y., Saroufim, C., Molloy, J., Wu, Xinyi, Arnold, S., Chang, Solomon, Schrittwieser, J., Buchatskaya, E., Radpour, S., Polacek, M., Giordano, S., Bapna, A., Tokumine, S., Hellendoorn, V., Sottiaux, T., Cogan, S., Severyn, A., Saleh, M., Thakoor, S., Shefey, L., Qiao, S., Gaba, M., Chang, Shuo-yiin, Swanson, C., Zhang, B., Lee, B., Rubenstein, P.K., Song, G., Kwiatkowski, T., Koop, A., Kannan, A., Kao, D., Schuh, P., Stjerngren, A., Ghiasi, G., Gibson, G., Vilnis, L., Yuan, Y., Ferreira, F.T., Kamath, A., Klimenko, T., Franko, K., Xiao, K., Bhattacharya, I., Patel, M., Wang, Rui, Morris, A., Strudel, R., Sharma, V., Choy, P., Hashemi, S.H., Landon, J., Finkelstein, M., Jhakra, P., Frye, J., Barnes, M., Mauger, M., Daun, D., Baatarsukh, K., Tung, M., Farhan, W., Michalewski, H., Viola, F., Quitry, F. de C., Lan, C. Le, Hudson, T., Wang, Qingze, Fischer, F., Zheng, I., White, E., Dragan, A., Alayrac, J., Ni, E., Pritzel, A., Iwanicki, A., Isard, M., Bulanova, A., Zilka, L., Dyer, E., Sachan, D., Srinivasan, S., Muckenhirn, H., Cai, H., Mandhane, A., Tariq, M., Rae, J.W., Wang, G., Ayoub, K., FitzGerald, N., Zhao, Y., Han, W., Alberti, C., Garrette, D., Krishnakumar, K., Gimenez, M., Levskaya, A., Sohn, D., Matak, J., Iturrate, I., Chang, M.B., Xiang, J., Cao, Y., Ranka, N., Brown, G., Hutter, A., Mirrokni, V., Chen, N., Yao, K.,

Egyed, Z., Galilee, F., Liechty, T., Kallakuri, P., Palmer, E., Ghemawat, S., Liu, Jasmine, Tao, D., Thornton, C., Green, T., Jasarevic, M., Lin, S., Cotruta, V., Tan, Y.-X., Fiedel, N., Yu, H., Chi, E., Neitz, A., Heitkaemper, J., Sinha, Anu, Zhou, D., Sun, Yi, Kaed, C., Hulse, B., Mishra, S., Georgaki, M., Kudugunta, S., Farabet, C., Shafran, I., Vlasic, D., Tsitsulin, A., Ananthanarayanan, R., Carin, A., Su, G., Sun, P., V, S., Carvajal, G., Broder, J., Comsa, I., Repina, A., Wong, W., Chen, W.W., Hawkins, P., Filonov, E., Loher, L., Hirnschall, C., Wang, W., Ye, J., Burns, A., Cate, H., Wright, D.G., Piccinini, F., Zhang, L., Lin, C.-C., Gog, I., Kulizhskaya, Y., Sreevatsa, A., Song, S., Cobo, L.C., Iyer, A., Tekur, C., Garrido, G., Xiao, Z., Kemp, R., Zheng, H.S., Li, H., Agarwal, A., Ngani, C., Goshvadi, K., Santamaria-Fernandez, R., Fica, W., Chen, Xinyun, Gorgolewski, C., Sun, S., Garg, R., Ye, X., Eslami, S.M.A., Hua, N., Simon, J., Joshi, P., Kim, Y., Tenney, I., Potluri, S., Thiet, L.N., Yuan, Q., Luisier, F., Chronopoulou, A., Scellato, S., Srinivasan, P., Chen, Minmin, Koverkathu, V., Dalibard, V., Xu, Yaming, Saeta, B., Anderson, K., Sellam, T., Fernando, N., Huot, F., Jung, J., Varadarajan, M., Quinn, M., Raul, A., Le, M., Habalov, R., Clark, J., Jalan, K., Bullard, K., Singhal, A., Luong, T., Wang, B., Rajayogam, S., Eisenschlos, J., Jia, J., Finchelstein, D., Yakubovich, A., Balle, D., Fink, M., Agarwal, S., Li, J., Dvijotham, D., Pal, S., Kang, K., Konzelmann, J., Beattie, J., Dousse, O., Wu, D., Crocker, R., Elkind, C., Jonnalagadda, S.R., Lee, Jong, Holtmann-Rice, D., Kallarackal, K., Liu, Rosanne, Vnukov, D., Vats, N., Invernizzi, L., Jafari, M., Zhou, H., Taylor, L., Prendki, J., Wu, M., Eccles, T., Liu, T., Kopparapu, K., Beaufays, F., Angermueller, C., Marzoca, A., Sarcar, S., Dib, H., Stanway, J., Perbet, F., Trdin, N., Sterneck, R., Khorlin, A., Li, Dinghua, Wu, Xihui, Goenka, S., Madras, D., Goldshtein, S., Gierke, W., Zhou, T., Liu, Yaxin, Liang, Y., White, A., Li, Yunjie, Singh, Shreya, Bahargam, S., Epstein, M., Basu, S., Lao, L., Ozturel, A., Crous, C., Zhai, A., Lu, H., Tung, Z., Gaur, N., Walton, A., Dixon, L., Zhang, M., Globerson, A., Uy, G., Bolt, A., Wiles, O., Nasr, M., Shumailov, I., Selvi, M., Piccinno, F., Aguilar, R., McCarthy, S., Khalman, M., Shukla, M., Galic, V., Carpenter, J., Villela, K., Zhang, H., Richardson, H., Martens, J., Bosnjak, M., Belle, S.R., Seibert, J., Alnahlawi, M., McWilliams, B., Singh, Sankalp, Louis, A., Ding, W., Popovici, D., Simicich, L., Knight, L., Mehta, P., Gupta, N., Shi, C., Fatehi, S., Mitrovic, J., Grills, A., Pagadora, J., Munkhdalai, T., Petrova, D., Eisenbud, D., Zhang, Zhishuai, Yates, D., Mittal, B., Tripuraneni, N., Assael, Y., Brovelli, T., Jain, P., Velimirovic, M., Akbulut, C., Mu, J., Macherey, W., Kumar, R., Xu, J., Qureshi, H., Comanici, G., Wiesner, J., Gong, Z., Ruddock, A., Bauer, M., Felt, N., GP, A., Arnab, A., Zelle, D., Rothfuss, J., Rosgen, B., Shenoy, A., Seybold, B., Li, Xinjian, Mudigonda, J., Erdogan, G., Xia, J., Simsa, J., Michi, A., Yao, Y., Yew, C., Kan, S., Caswell, I., Radebaugh, C., Elisseeff, A., Valenzuela, P., McKinney, K., Paterson, K., Cui, A., Latorre-Chimoto, E., Kim, S., Zeng, W., Durden, K., Ponnapalli, P., Sosea, T., Choquette-Choo, C.A., Manyika, J., Robenek, B., Vashisht, H., Pereira, S., Lam, H., Velic, M., Owusu-Afriyie, D., Lee, K., Bolukbasi, T., Parrish, A., Lu, S., Park, J., Venkatraman, B., Talbert, A., Rosique, L., Cheng, Yuchung, Sozanschi, A., Paszke, A., Kumar, P., Austin, Jessica, Li, L., Salama, K., Perz, B., Kim, W., Dukkipati, N., Baryshnikov, A., Kaplanis, C., Sheng, X., Chervonyi, Y., Unlu, C., Casas, D. de Las, Askham, H., Tunyasuvunakool, K., Gimeno, F., Poder, S., Kwak, C., Miecnikowski, M., Mirrokni, V., Dimitriev, A., Parisi, A., Liu, D., Tsai, T., Shevlane, T., Kouridi, C., Garmon, D., Goedeckemeyer, A., Brown, A.R., Vijayakumar, A., Elqursh, A., Jazayeri, S., Huang, J., Carthy, S.M., Hoover, J., Kim, L., Kumar, S., Chen, W., Biles, C., Bingham, G., Rosen, E., Wang, Lisa, Tan, Q., Engel, D., Pongetti, F., de Cesare, D., Hwang, D., Yu, L., Pullman, J., Narayanan, S., Levin, K., Gopal, S., Li, M., Aharoni, A., Trinh, T., Lo, J., Casagrande, N., Vij, R., Matthey, L., Ramadhana, B., Matthews, A., Carey, C., Johnson, Matthew, Goranova, K., Shah, R., Ashraf, S., Dasgupta, K., Larsen, R., Wang, Y., Vuyyuru, M.R., Jiang, C., Ijazi, J., Osawa, K., Smith, Celine, Boppana, R.S., Bilal, T., Koizumi, Y., Xu, Ying, Altun, Y., Shabat, N., Bariach, B., Korchemniy, A., Choo, K., Ronneberger, O., Iwuanyanwu, C., Zhao, S., Soergel, D., Hsieh, C.-J., Cai, I., Iqbal, S., Sundermeyer, M., Chen, Z., Bursztein, E., Malaviya,